\documentclass{article}
\usepackage{geometry}
\usepackage{ijcai22}

\usepackage{times}

\usepackage{soul}
\usepackage{url}
\usepackage[hidelinks]{hyperref}
\usepackage[utf8]{inputenc}
\usepackage[small]{caption}
\usepackage{graphicx}
\usepackage{amsmath}
\usepackage{booktabs}
\usepackage{float}

\usepackage{cuted}
\usepackage{flushend}
\usepackage{times}
\usepackage{soul}
\usepackage{url}
\usepackage{multirow}
\usepackage{geometry}
\usepackage{graphicx}
\usepackage{algorithm}
\usepackage{algorithmic}
\usepackage{amsthm}
\usepackage{subfigure}
\usepackage{amssymb}
\usepackage[misc]{ifsym}
\usepackage{multicol}
\usepackage{amsmath}
\usepackage{xcolor}
\newtheorem{theorem}{Theorem}
\newtheorem{lemma}{Lemma}

\usepackage[hidelinks]{hyperref}
\usepackage[utf8]{inputenc}
\usepackage{booktabs}
\usepackage[normalem]{ulem}
\usepackage{multirow}
\title{Exploit Customer Life-time Value with Memoryless Experiments}

\author{
Zizhao Zhang$^1$\footnote{Three authors contributed equally to this research.}
\and
Yifei Zhao$^1$\footnotemark[1]
\And
Guangda Huzhang$^1$\footnotemark[1]
\footnote{Contact Author.}\\
\affiliations
$^1$Alibaba Group\\
\emails
zhangzizhao.zzz@alibaba-inc.com,
andy.zyf@alibaba-inc.com,
guangda.hzgd@alibaba-inc.com
}

\begin{document}

\maketitle

\begin{abstract}
As a measure of the long-term contribution produced by customers in a service or product relationship, life-time value, or LTV, can more comprehensively find the optimal strategy for service delivery. However, it is challenging to accurately abstract the LTV scene, model it reasonably, and find the optimal solution. The current theories either cannot precisely express LTV because of the single modeling structure, or there is no efficient solution. We propose a general LTV modeling method, which solves the problem that customers' long-term contribution is difficult to quantify while existing methods, such as modeling the click-through rate, only pursue the short-term contribution. At the same time, we also propose a fast dynamic programming solution based on a mutated bisection method and the memoryless repeated experiments assumption. The model and method can be applied to different service scenarios, such as the recommendation system. Experiments on real-world datasets confirm the effectiveness of the proposed model and optimization method. In addition, this whole LTV structure was deployed at a large E-commerce mobile phone application, where it managed to select optimal push message sending time and achieved a 10\% LTV improvement. 

\end{abstract}

\section{Introduction}

Life-time value (LTV), which means the total value a customer could contribute to a specific business or program over the whole period of their interactions, anticipates how much profit this business or program is consistently making or will make. It plays an increasingly essential role in today's internet companies for driving business growth \cite{zeithaml2001driving,paquienseguy2018user,shiau2018examining}. Different from referring to some traditional product measurements such as click-through rate (CTR), whose goal is to maximize engagement of customers or to earn profit such as the advertising revenue in a short period, LTV also emphasizes to collect the long-term value a customer could produce from the current product or service while the others risk to an immediate user churn. For instance, as one of the current mainstream recommendation systems, the sequential recommendation can display almost an unlimited number of advertisements to a customer \cite{chen2018sequential,huang2018improving}. There are two main objectives that sequential recommendation aims to achieve, attracting customers to browse as many items as possible and capturing the interests of customers as accurately as possible. 
However, the only pursuit of CTR may harm the 
former measurement, and end with a sub-optimal solution because of the conflict between a longer browsing session and a higher immediate customer engagement. From a long-run perspective, LTV runs over CTR in such scenarios and can produce a significantly higher revenue. \cite{kumar2004customer,chan2011measuring}.

To maximize the LTV of a product or service is not a trivial task. The Pareto/NBD framework is the first model to address LTV analysis, but it has been proved that it is hard to be applied in the real world during these years. In many circumstances, modeling the LTV structure usually means incorporating a lot of stochastic complexity, which makes the optimization extremely challenging. For instance, \cite{permana2014study} applied the Markov chain model but admitted a very long convergence time. 

In this paper, we propose a general life-time value optimization framework that is suitable for the scenario that decisions are made in discrete periods. 
Our framework can be regarded as an infinite round of independent experiments with the memoryless assumption on customers.
The process can be described as directed acyclic graphs to represent various states and their transitions, and the life-time value can be efficiently maximized. We name our framework memoryless repeated experiments optimization (MREOpt). The high-level idea of MREOpt is to apply a mutated bisection method to solve a general-form dynamic programming problem. We use MREOpt to decide the optimal push message sending time on an anonymous E-commerce mobile phone application (App). Our goal is to attract customers to click these push messages as many times as possible while minimizing the chance they turn off the message channel or refuse to receive the App notifications anymore. Results of our experiments show the proposed method significantly outperforms the previous short-sighted method. The offline experiments show an up-to 50\% LTV improvement of the proposed model  and an online A/B test also confirms its effectiveness by generating 10\% more real business revenue.

Our main contributions are
\begin{itemize}
    \item We propose a solid modeling method for LTV. This method is based on the memoryless repeated experiments assumption and captures not only customers' short-term rewards, content clicks, or engagements, for instance, but also makes customers engage with the service or product as long as possible. 
    \item Given this novel LTV modeling structure, we propose a dynamic programming algorithm to optimize the LTV. This algorithm incorporates a mutated bisection method as the key ingredient to efficiently optimize LTV under a dynamic programming structure.
    \item This whole LTV modeling and solution system can be applied in scenarios where LTV plays as the main interest. Its generality and expandability can easily add other related variables, such as time, customers' features, etc., into the modeling process.
    \item The proposed model is deployed on an E-commerce mobile phone application for deciding the optimal push message sending time for customers and achieves a 10\% enhancement on revenue.
\end{itemize}

The rest of the paper is organized as follows. In section 2, we discuss related works. After providing fundamental model definitions in section 3, section 4 provides the LTV optimization problem, different modeling strategies given various state transitions, and the proposed algorithm. Experiments results are shown in section 5, and section 6 concludes the paper. 

\section{Related Work}

Although the concept of customer life-time value has been defined for decades, there is not much related work, especially from a mathematical modeling perspective. To our best acknowledgement, some recent modeling developments include applying decision trees \cite{bin2007customer,ma2009research,kirui2013predicting}, neural network \cite{hadden2007computer,glady2009modeling,karahoca2011gsm}, Markov chain model \cite{burez2009handling,dierkes2011estimating} and association rules \cite{tsai2010variable}. One main problem of these works is that they rarely plug customer behavior into modeling and lack of future value estimation.  

Optimizing LTV can be formulated as the stochastic shortest path (SSP) problem, and a recent work is from \cite{zhao2020maximizing}, where a similar goal is proposed to improve the sequential recommendation. However, the SSP structure usually neglects the long-term contribution or value.  The fundamental behind the SSP is the Markov decision process, and some related applications can be found in \cite{barto1995learning,kolobov2011heuristic}.

\section{Preliminary}
We describe the process of sending push messages to a customer's mobile phone in the Markov Decision Process form.
\subsection{Markov Decision Process}
A  Markov decision process (MDP) 4-tuple $(S, A, P, R)$ can be expressed by 
\begin{itemize}
    \item $S$ is a set of states $s_i \in S, \; i= 1,\cdots$, in our case, state set $S$ depicts users' different reactions after receiving a push message, including continue allowing to receive more pushes and turn off the push channel; 
    \item $A$ is a set of actions $a_i \in A, \; i = 1, \cdots$, action set $A$ includes two actions: the App keeps sending or stops sending the push to the customer;
    \item $P_{s,a,s'}  = \mathbf{P}(s_{t+1}= s' | s_t = s, a_t = a)$ is the probability that action $a$ in state $s$ at time $t$ will lead to state $s'$ at time $t+1$, for each customer (we may use $p$ to denote a general probability);
    \item  $R_{s,a,s'} = R_{a}(s, s')$ is the immediate reward (or expected immediate reward) received after the transition from state $s$ to state $s'$, due to action $a$ (we may use $r$ to denote a general reward).

\end{itemize} 
 Note that given a state and an action, we have
$$\Sigma_{(s, s')} P_{s,a,s'} = 1, \; \forall s\in S,\forall a\in A.$$

\subsection{Stochastic Shortest Path Problem}
Based on an MDP model, the optimization of LTV can be formulated as an SSP problem, whose objective is to find a strategy $\pi$: $S \rightarrow A$ that maximizes a customer's long-term contribution, i.e.,
\begin{equation}
    \pi = \text{argmax}_{\pi} \mathbf{E}(\sum^{T}_{t=1} R_{s_t, a_t}),
\end{equation}
where $T$ is the life length. The distribution of $T$ is 
\begin{equation}
    \mathbf{P}(T > t) = \prod_{i<t}P_{s_i, a_i, s_{i+1}}.
\end{equation}
Combining the above two equations, we have 
\begin{equation}
\label{equation: objective}
    \mathbf{E}(\sum^{T}_{t=1} R_{s_t, a_t}) = \sum^{T}_{t=1}R_{s_t,a_t}\times \prod_{i<t}P_{s_i, a_i, s_{i+1}}.
\end{equation}

Therefore, to optimize equation (\ref{equation: objective}), we need to estimate each $R_{s_t,a_t}$ and $P_{s_i,a_i,s_A}$.

\section{Modeling and Methods}

Our objective is to select optimal push message sending time for customers on a large $E$-commerce App, with which we achieve these customers' maximal LTV. We illustrate this process under an MDP framework. The most important step in an MDP is to determine its state transition.

\subsection{State Transition Modeling}

We start with introducing a fundamental state transition structure, where the sending time is not considered. After we send a push to a customer, there are two states: \textit{(i).} we can continue to send one more push; \textit{(ii).} the customer turns off the push channel, and we conclude the customer is lost. It is obvious to note that if such state sequence is finite, then we are facing a classic stochastic shortest path  problem, which tries to find the path from the initial state to the end state with the highest expected return value. The SSP problem can be solved by recursion \cite{ji2005models}. Working with this transition structure cannot select a more appropriate sending time to access the customer. As a proactive customer reaching channel, sending push messages can easily cross the line and make customers feel disturbed, which happens more often if sending time is bad, say, when the customer is having a class or a meeting. Hence, putting the sending time into consideration should be necessary. 

\begin{figure}[t]
\includegraphics[width=7.8cm]{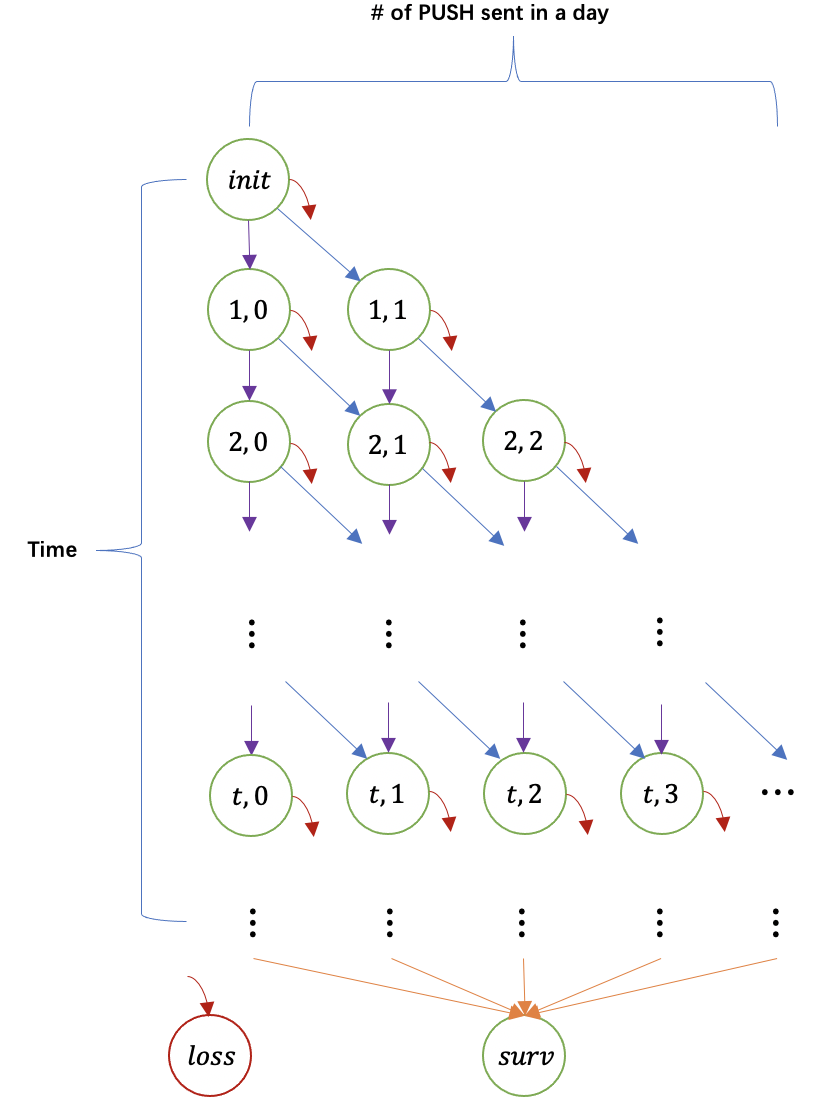}
\caption{The number of push messages sent in a day given a list of sending time candidates and each possible state/action a customer may take.}
\label{fig: state transition 2}
\centering
\end{figure}

We extend the above state transition structure by incorporating time factor $t$. Obviously, we require the sending time of the $i^{th}$ push is ahead of the $(i+1)^{th}$ push. The system structure is shown in  figure \ref{fig: state transition 2}, where the $(t, m)^{th}$ cell represents a state that in the first $t$ time units, we send $m$ pushes to the customer. At every state with time $t$, two actions can be chosen and implemented: send or not send push. If the former, we follow the blue arrow and transfer the state from $(t, m)$ to $(t+1, m+1)$; and if the latter, the state follows the purple arrow and reaches $(t+1, m)$. In addition, a red arrow exists, meaning that at every state, the customer might be lost because of our action. Assuming the state sequence is finite, the whole structure is, in essence, a directed acyclic graph (DAG). The optimal strategy can be deducted by reversing its topological order \cite{handley1994use}.

In fact, this structure with time $t$ describes a customer's state transition within a given day. Since we are dealing with a long-run process, it should be expanded. Here we define a directed graph $G = (V, E)$ as follows
\begin{equation}
\begin{aligned}
\nonumber 
V & = \{s_{init}, s_{loss}, s_{surv}, s_1, s_2, s_3, \cdots \},\\
E & = \{(s_{init}, s_i)\} \cup \{(s_i, s_j)\}  \\ 
&   \cup \{(s_i, s_{loss})\} \cup \{(s_i, s_{surv})\},
\end{aligned}
\end{equation}
where $(s_{init}, s_i)$ and $(s_i, s_j)$ indicate an effective state transition, $(s_i, s_{loss})$ represents the action makes the customer leave and $(s_i, s_{surv})$ means the customer finally survives.

Unlike the standard type of dynamic programming problems, this graph problem contains either an infinite number of states with no loop transition, or finite states with loop transition. Whichever type it is, it is hard to solve it as a Markov decision process problem.

To effectively simulate our scenario, we propose to consider using the memoryless repeated experiments assumption (MREA), whose practical value has been highly appreciated in many fields \cite{jerath2011new}. One famous application of MREA in the economy is the ``Buy Till You Die'' model \cite{fader2005counting}. A key point of MREA is that if a customer keeps active to the next period, the future value he would create is the same as the value he creates since this period.  The initial state of a customer on day one is exactly the same as day two and so. The strategy can be interpreted as we keep sending pushes to a customer until he leaves this channel. Mathematically, following MREA, we have 
$$f(s_{init}^1) = f(s_{init}^2) = \cdots = f(s_{surv}),$$
where $s^i_{init}$ is the initial state on day $i$.

\begin{figure}
\includegraphics[width=8cm]{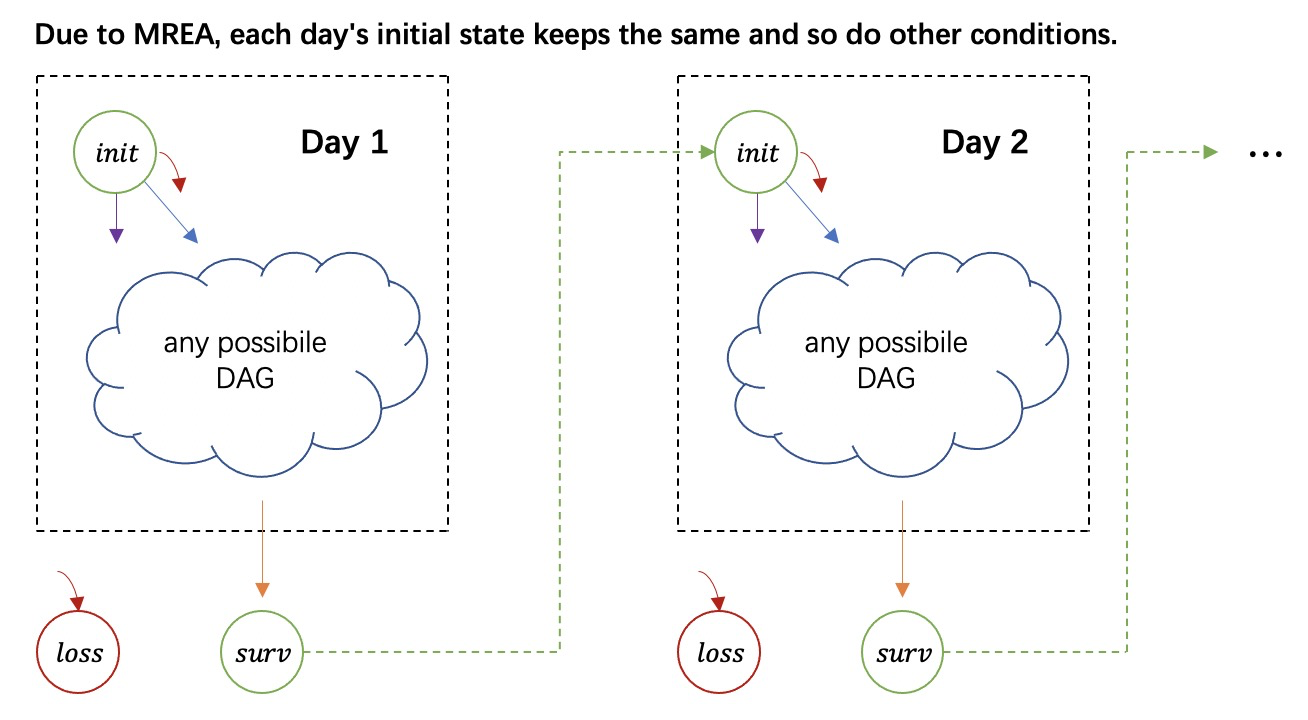}
\caption{The diagram of state transition under MREA.}
\label{fig: state transition 3}
\centering
\end{figure}

The above state transition structure is displayed in figure \ref{fig: state transition 3}. If we define 
$$DAG = G\{g_1, g_2, \cdots\},$$
where $DAG$ represents a network structure described by a high-dimensional functional $G$. This functional $G$ can be easily extended by including other factors, besides time $t$ in our push sending case. For instance, when we decide to include item recommendation function, the push sending $DAG$ is now written as 
$$DAG = G\{g_1(\text{time}), g_2(\text{item})\},$$
and the network is expanded to two-dimensional.

Let $\pi : S \rightarrow A$ denote a strategy and trajectory $\tau_\pi = (s_1, a_1, r_1, s_2, a_2, r_2, ...)$ sampled from $\pi$. To optimize $f(s_{surv})$, we are supposed to solve
\begin{equation}
    \text{argmax}_{\pi} \mathbf{E}_{\tau_\pi \sim \pi}[\sum^{\infty}_{i=1}r_i].
    \label{equation: 1}
\end{equation}

Since the trajectory in problem (\ref{equation: 1}) may be infinitely long, it cannot be directly optimized.

Based on our previous illustration, optimizing function (1) is equal to finding the optimal value at each period. 

\subsection{Optimization Methods}
\subsubsection{Brute Force}
Consider the 1-day case, we need to optimize
\begin{equation}
    f(s) = \max_{a\in A} \sum_{s'} P_{s, a, s'}\cdot [R_{s, a, s'} + f(s')],
\end{equation}
with the initial condition 
\begin{equation}
    f(s_{loss}) = f(s_{surv}) = 0.
\end{equation}

The corresponding brute force method is shown in algorithm \ref{alg: ego}. After expanding figure \ref{fig: state transition 3} into an appropriate scale, say, day 1 to day 100, the above brute force method is able to solve the problem, and we can use the solution to approximate the exact one. However, there are two drawbacks. First, the approximated solution might have a significant gap to the exact one; second, the computational complexity would be enlarged by multiplying a large constant.  

\newfloat{algorithm}{!ht}{lop}
\begin{algorithm}
\caption{One-round Value Optimization (BF)}
\label{alg: ego}
\begin{algorithmic}
\STATE {\textbf{Input: } transition probability $P_{s_i, a_j, s_k} \in [0, 1]$ and reward $R_{s_i, a_j, s_k} \in [0, 1]$ if we apply action $a_j\in A$ on state $s_i \in S$ and transition to state $s_k \in S$.}
\STATE {\textbf{Output: } the optimized one-round value.}
\STATE{\textbf{Initialization:} $N_s \gets |S|$, $N_a \gets |A|$,}
\STATE{ $f(\cdot) \gets - \infty$,$f(s_{surv})\gets 0$, $f(s_{loss})\gets 0$.}
\STATE{Compute the inversed topologic order $L=(s_{surv}, s_{loss}, s_1, s_2, ..., s_{N_s}, s_{init})$} 
\FOR {each $s \in L \setminus \{s_{surv},s_{loss}\} $} 
\FOR {each $a \in A$} 
\STATE {$v_a = 0$}
\FOR {each $s' \in S$ where $P_{s, a, s'} > 0$}
\STATE {$v_a \gets v_a + P_{s, a, s'} \cdot ( R_{s, a, s'} + f(s'))$}
\ENDFOR 
\STATE {$f(s) \gets \max(f(s), v_a)$}
\ENDFOR
\ENDFOR
\RETURN{$f(s_{init})$}
\end{algorithmic}
\end{algorithm}
\subsubsection{MREOpt}

\newfloat{algorithm}{!htb}{lop}
\begin{algorithm}
\caption{Infinite-rounds Value Optimization (MREOpt)}
\label{alg: ego 2}
\begin{algorithmic}
\STATE {\textbf{Input: } transition matrix $P_{s_i, a_j, s_k} \in [0, 1]$ and reward $R_{s_i, a_j, s_k} \in [0, 1]$ if we apply action $a_j\in A$ on state $s_i \in S$ and transition to state $s_k \in S$.}
\STATE {\textbf{Output: } the optimized life-time value.}
\STATE{\textbf{Initialization: } $N_s \gets |S|$, $N_a \gets |A|$,}
\STATE{ $f(\cdot) \gets - \infty$, $\hat f(s_{surv})\gets (1, 0)$, }
\STATE{$\hat f(s_{loss})\gets (0, 0)$, $left \gets 0$, $right \gets \infty $.}
\STATE{Compute inversed topologic order $L=(s_{surv}, s_{loss}, s_1, s_2, ..., s_{N_s}, s_{init})$} 
\WHILE {$right - left > \epsilon$, e.g., $\epsilon\gets 0.01$}
\STATE{$g \gets \frac{left + right}{2}$}
\FOR {each $s \in L \setminus \{s_{surv},s_{loss}\} $} 
\FOR {each $a \in A$} 
\STATE {$p_a \gets 0, r_a \gets 0$}
\FOR {each $s' \in S$ where $P_{s, a, s'} > 0$}
\STATE {$p_a \gets p_a + P_{s, a, s'} \cdot \hat f(s').p$}
\STATE {$r_a \gets r_a + P_{s, a, s'} \cdot ( R_{s, a, s'}  + \hat f(s'))$}
\ENDFOR 
\IF {$\hat f(s). p \cdot g + \hat f(s). r < p_a \cdot g + r_a$}
  \STATE {$\hat f(s). p \gets p_a$, $\hat f(s).   r \gets r_a$}
\ENDIF
\ENDFOR
\ENDFOR
\IF {$\frac{\hat f(s_{init}). r}{1-\hat f(s_{init}). p} \leq g$}
  \STATE {$left\gets g$}
\ELSE
  \STATE {$right\gets g$}
\ENDIF
\ENDWHILE
\RETURN{$left$}
\end{algorithmic}
\end{algorithm}

In this subsection, we first clearly distinguish short and long-term LTV that appeared in problem. We know that $f(s_{init})$ would eventually transfer to $f(s_{surv})$ with probability $p$, or we get the customer lost after receiving a short-term return $r$ with probability $1-p$. In our case, $p$ is the probability that a customer clicks the push. Due to MREA, for the long-term LTV, we have $f(s_{init}) = f(s_{surv})$ and 
\begin{equation}
\label{equation: 4}
    f(s_{init}) = pf(s_{surv}) + r = pf(s_{init}) + r,
\end{equation}
thus 
\begin{equation}
\label{equation: 5}
    f(s_{init}) = \frac{r}{1-p}.
\end{equation}

Given equations (\ref{equation: 4}) and (\ref{equation: 5}), we factorize $f(s_i)$ by referring to the short and long-term LTV. For convenience, we define
$$\hat{f}_{s_i} := (\hat{f}(s_i). p, \hat{f}(s_i). r).$$
This quantity represents that when period $i$ ends, a customer can generate a long-term LTV with probability $p$ and obtain a short-term return $r$. 

The only deterministic final state has to be defined accordingly. We thus have $\hat{f}(s_{loss}) = (0, 0)$ and $\hat{f}(s_{surv}) = (1, 0)$. 

We propose a dynamic programming method, namely memoryless repeated experiments optimization (MREOpt), to maximize LTV given the unlimited rounds of state, and the memoryless repeated experiments assumption. We begin searching the best state transition trajectory by an initial guess value and then iteratively use the bisection method to optimize the return value. The algorithm detail is shown in algorithm \ref{alg: ego 2}.

The main difference between algorithm \ref{alg: ego} and \ref{alg: ego 2} lies in the change of recursion equations ,and the latter one is  

\begin{equation}
\nonumber 
    \begin{split}
  &  \qquad    (\hat{f}(s). p, \hat{f}(s) . r) \\
 & =  \max_{a\in A}^{g} \{\sum_{s'}P_{s,a,s'}\cdot f(s'). p, \\ 
 & \quad \sum_{s'}P_{s,a,s'}\cdot [R_{s,a,s'}+f(s'). r]\}.
    \end{split}
\end{equation}

Note here we define ``max'' operation with respect to $g$ by the rule
$$(p, r) < (p', r') \Leftrightarrow p\cdot g + r < p'\cdot g + r',$$
which expresses that the best short and long-term combination is selected given the long-term value is $g$. More specifically, if a strategy can earn a short-term return $r$ with probability $p$ being alive, it would continue to earn a long-term return $g$. This rule maps a combination of $(r, g)$ to a singular value, which can be solved by a usual recurrence relation.

We then prove the correctness of algorithm \ref{alg: ego 2}.

{\lemma If $g$ is monotonically increasing, $\hat{f}(s_{init}) . p \cdot g + \hat{f}(s_{init}). r$ is increasing and continuous. }

\begin{proof}
 If the optimal strategy does not change when $g$ increases, then $\hat{f}(s_{init}). p$ and $\hat{f}(s_{init}). r$ does not change. If the optimal strategy is about to change, which means there exists a pair $(p', r')$ such that $p\cdot g + r = p'\cdot g + r'$, we switch their values. If we use $(p^\ast, r^\ast)$ to denote the survival probability and short-term LTV at the changing state point, the two strategies satisfy 
$$p\cdot g + r =  p'\cdot g + r' $$ 
$$ \Rightarrow p^\ast\cdot (p\cdot g + r) + r^\ast = p^\ast \cdot (p'\cdot g + r') + r^\ast.$$
Therefore, the strategy switching does not change the result. Function $\hat{f}(s_{init}) . p \cdot g + \hat{f}(s_{init}). r$ is increasing and continuous. 
\end{proof}

{\lemma If we define $F(g) = \hat{f}(s_{init}). p \cdot g + \hat{f}(s_{init}). r$, then $F$ is continuous and there exists $g'$ satisfying $$g' = \frac{\hat{f}(s_{init}). r}{1 - \hat{f}(s_{init}). p}.$$}

\begin{proof}
Based on the previous context, we know that $\frac{\hat{f}(s_{init}). r}{1 - \hat{f}(s_{init}). p}$ is the actual LTV generated by the optimal strategy, which is 
\begin{equation}
\begin{aligned}
& \qquad   \frac{\hat{f}(s_{init}). r}{1 - \hat{f}(s_{init}). p}  \\ 
& =    \hat{f}(s_{init}) . r + \hat{f}(s_{init}). p \cdot \hat{f}(s_{init}). r \\ 
& + \hat{f}(s_{init}). p^{2} \cdot \hat{f}(s_{init}). r + \cdots.
\end{aligned}
\end{equation}

As a guessed LTV, function $g$ only affects the strategy selection during state transition. When $g = 0$, $F(g) = \hat{f}(s_{init}) . r \geq 0$, so $g \leq F(g)$; when $g \rightarrow \infty$, since $p, r \leq 1$, we have $F(g) = \hat{f}(s_{init}). p \cdot g + \hat{f}(s_{init}. r)\leq g$. From lemma 1 and the mean value theorem, there exists a $g'$ satisfying $F(g) = g'$, so 
$$F(g) = \hat{f}(s_{init}). p \cdot g' + \hat{f}(s_{init}). r = g' $$ 
$$ \Rightarrow g' = \frac{\hat{f}(s_{init}). r}{1 - \hat{f}(s_{init}. p)}.$$
\end{proof}

{\theorem  Algorithm \ref{alg: ego 2} can optimize the customer life-time value offline and find the best combination of actions.}

\begin{proof}
From lemma 1 and lemma 2, there is exactly one interval $\Omega$ satisfying $F(g) = g'$. We can start the iteration by picking a strategy $g^\dagger \in \Omega^{C}$. On the one hand, algorithm \ref{alg: ego 2} guarantees to update $g^\dagger$ to be more close to $\Omega$; and on the other hand, the set of optional strategies is finite. Therefore, algorithm \ref{alg: ego 2} maximizes LTV.

If a customer is not lost or leaves the channel, his feedback does not cause influence on the optimal strategy; otherwise, a lost customer cannot generate subsequent behaviors. Thus algorithm \ref{alg: ego 2} can work offline.
\end{proof}
 
\section{Experiments and Results}
In this section, based on offline historical data, we test whether the LTV model proposed in the last section can achieve a higher LTV compared to other methods. The experiments in this section are implemented on a large E-commerce App.

\subsection{Dataset and Prediction Model}
\textbf{Dataset:} The dataset comes from the E-commerce App and is about push message sending records of this App. The size of this dataset is up to ten billion.

Based on the dataset, the App already obtains a well-trained model that can provide the probability of a customer clicking a specific push message given the customer's personal information and the push content. Moreover, it also has an accurate model that calculates the probability  of a customer closing the push message channel. The first model can work as the approach to learning the reward function in the push engagement process,  while the second model can provide information implying whether a customer is about to quit or end the service.

\subsection{Evaluation Measures}

In the experiment, we consider the cumulative clicks as cumulative customer engagement. Without loss of generality, we name cumulative clicks as \textit{LTV} for short. Moreover, we use  \textit{LT} to indicate life-time. In offline evaluation, suppose that the number of push message sequences is $T$. These two indicators are  defined as  
\begin{equation}
    LTV = \sum^T_{t=1} R_{s_t, a_t} \times \prod_{i<t} (1 - P_{s_i, a_i, s_A}),
\end{equation}
\begin{equation}
    LT = \sum^T_{t=1}\prod_{i<t}(1-P_{s_i,a_i,A}).
\end{equation}

We define \textit{CTR} of the push sending sequence as 
\begin{equation}
    CTR = \frac{LTV}{LT}.
\end{equation}

In online evaluation, \textit{LTV} can be counted as 
\begin{equation}
    LTV = \sum^{T}_{t=1}c_t,
\end{equation}
where $c_t \in \{0, 1\}$ is the click behavior at step $t$, and $T$ is the life-time, i.e., $LT = T$.

In the following parts, when we report \textit{LTV}, \textit{LT} and \textit{CTR}, they are referring to the corresponding average values summarized over ten million randomly selected customers from the customer pool.

\subsection{Compared Strategies}

In the experiment, we compare MREOpt with the greedy strategy and the BF method. The main characteristic of the greedy strategy is the customers' quitting probability, or push channel closing probability, is not considered, and it only tries to find a path maximizing the clicking rates. As we introduced earlier, the BF method can be used to optimize LTV as long as the sequence is finite.

In the experiment, we assume the App can send a large number of push messages to a customer in a day, which is, though, not likely in reality. Each sending time comes from a candidate time pool whose size is denoted as $\Lambda$. We use the notation $M$ to denote the maximum number of push messages that we can send to a customer in a day.

\subsection{Results and Analysis}

The detailed offline experiment results are shown in table \ref{table: results 1} to \ref{table: results 3}, and we can find that  the greedy method has the best \textit{CTR} while MREOpt achieves the best \textit{LTV} and \textit{LT}. The greedy method does not care about the customer's life span but aggressively picks the best move to gain immediate reward, i.e., higher click rate, without considering customers' quitting probability. As we expect, the MREOpt method plans to improve \textit{LTV} through making customers stay longer in the service. Although the brute force method also includes the quitting probability into the decision process, it, as we state, loses its efficiency when the state sequence is very long or close to infinity. 

 \begin{table}[!h]
     \centering
\resizebox{.48\textwidth}{!}{
     \begin{tabular}{ccccccc} \toprule
   \multirow{ 2}{*}{Method}   & \multicolumn{3}{c}{$M=10$} &   \multicolumn{3}{c}{$M=50$} \\
      & \textit{LTV} & \textit{LT} & \textit{CTR}  & \textit{LTV} & \textit{LT} & \textit{CTR} \\ \midrule 
    Greedy& 0.609  &  1.499  &  0.407  &  0.697  &  1.763  &  0.395   \\
     BF & 0.781  &  2.063  &  0.378  &  0.724  &  2.116  &  0.342   \\
    MREOpt & 0.817  &  2.239  &  0.365  &  0.792  &  2.139  &  0.37  \\  \bottomrule 
     \end{tabular}
     }
     \caption{Offline evaluation results when $\Lambda = 100$.}
     \label{table: results 1}
 \end{table}

 \begin{table}[!h]
     \centering
\resizebox{.48\textwidth}{!}{
       \begin{tabular}{ccccccc} \toprule
   \multirow{ 2}{*}{Method}   & \multicolumn{3}{c}{$M=10$} &   \multicolumn{3}{c}{$M=50$} \\
      & \textit{LTV} & \textit{LT} & \textit{CTR}  & \textit{LTV} & \textit{LT} & \textit{CTR} \\ \midrule  
       Greedy & 0.547  &  1.346  &  0.406  &  0.613  &  1.513  &  0.405    \\
        BF & 0.793  &  2.078  &  0.382  &  0.756  &  2.169  &  0.349    \\
     MREOpt & 0.821  &  2.296  &  0.358  &  0.824  &  2.315  &  0.356   \\ \bottomrule 
     \end{tabular}}
     \caption{Offline evaluation results when $\Lambda = 200$.}
     \label{table: results 2}
 \end{table}

 \begin{table}[!h]
     \centering
\resizebox{.48\textwidth}{!}{
       \begin{tabular}{ccccccc} \toprule 
   \multirow{ 2}{*}{Method}   & \multicolumn{3}{c}{$M=10$} &   \multicolumn{3}{c}{$M=50$} \\
      & \textit{LTV} & \textit{LT} & \textit{CTR}  & \textit{LTV} & \textit{LT} & \textit{CTR} \\ \midrule 
       Greedy & 0.588 &  1.434  &  0.410  &  0.669  &  1.642  &  0.407    \\
        BF & 0.800  &  2.091  &  0.383  &  0.763  &  2.172  &  0.351    \\
     MREOpt & 0.846  &  2.422  &  0.349  &  0.851  &  2.435  &  0.349  \\ 
   \bottomrule 
     \end{tabular}}
     \caption{Offline evaluation results when $\Lambda = 500$.}
     \label{table: results 3}
 \end{table}

\subsection{Online Evaluation}
After the offline evaluation confirming the effectiveness of the model and the MREOpt algorithm, we deploy this system on a real E-commerce App. A one-month A/B test is conducted to see each strategy's online effects. Besides LTV, another direct online measurement is the customers' revenue contribution or gross merchandise value (GMV). Because of the data privacy protocol, we here do not exhibit exact business statistics but use the greedy method as the benchmark, and show the relative improvement of the other two algorithms.
\begin{table}[!h]
    \centering
    \begin{tabular}{ccccc} \toprule 
       Method  & \textit{LTV+\%} & \textit{LT+\%} &  \textit{GMV+\%} \\ \midrule 
       Greedy & -- & -- & -- \\ 
        BF & 6.2\% & 11.5\% & 2.8\%\\ 
        MREOpt & 10.1\% & 14.7\% & 5.1\%\\ \bottomrule 
    \end{tabular}
    \caption{Online evalution of \textit{LTV}, \textit{LT} and \textit{GMV} improvement when benchmark comes from the greedy method. }
    \label{table: online results}
\end{table}

From table \ref{table: online results}, we can see that our proposed algorithm MREOpt generates significantly higher \textit{LTV}, \text{LT} compared to other methodologies, which also produces real business value. 

\section{Summary}
This paper proposes a novel LTV modeling structure and an effective optimization algorithm MREOpt for maximizing customers' life-time value. This algorithm is built on the memoryless repeated experiments assumption and iteratively combines the bisection method and dynamic programming to update solutions. In our experiment, we manage to apply it to decide the optimal push message sending time on a large E-commerce App. Offline experiment results show that MREOpt can achieve an up to 60\% improvement optimizing customers' life-time value compared to commonly-used methods. Online results also support the superior advantage of MREOpt because it brings 5\% GMV enhancement.
\clearpage
\bibliographystyle{named}
\bibliography{ijcai22}

\begin{thebibliography}{}

\bibitem[\protect\citeauthoryear{Barto \bgroup \em et al.\egroup
  }{1995}]{barto1995learning}
Andrew~G Barto, Steven~J Bradtke, and Satinder~P Singh.
\newblock Learning to act using real-time dynamic programming.
\newblock {\em Artificial intelligence}, 72(1-2):81--138, 1995.

\bibitem[\protect\citeauthoryear{Bin \bgroup \em et al.\egroup
  }{2007}]{bin2007customer}
Luo Bin, Shao Peiji, and Liu Juan.
\newblock Customer churn prediction based on the decision tree in personal
  handyphone system service.
\newblock In {\em 2007 International Conference on Service Systems and Service
  Management}, pages 1--5. IEEE, 2007.

\bibitem[\protect\citeauthoryear{Burez and Van~den
  Poel}{2009}]{burez2009handling}
Jonathan Burez and Dirk Van~den Poel.
\newblock Handling class imbalance in customer churn prediction.
\newblock {\em Expert Systems with Applications}, 36(3):4626--4636, 2009.

\bibitem[\protect\citeauthoryear{Chan \bgroup \em et al.\egroup
  }{2011}]{chan2011measuring}
Tat~Y Chan, Chunhua Wu, and Ying Xie.
\newblock Measuring the lifetime value of customers acquired from google search
  advertising.
\newblock {\em Marketing Science}, 30(5):837--850, 2011.

\bibitem[\protect\citeauthoryear{Chen \bgroup \em et al.\egroup
  }{2018}]{chen2018sequential}
Xu~Chen, Hongteng Xu, Yongfeng Zhang, Jiaxi Tang, Yixin Cao, Zheng Qin, and
  Hongyuan Zha.
\newblock Sequential recommendation with user memory networks.
\newblock In {\em Proceedings of the eleventh ACM international conference on
  web search and data mining}, pages 108--116, 2018.

\bibitem[\protect\citeauthoryear{Dierkes \bgroup \em et al.\egroup
  }{2011}]{dierkes2011estimating}
Torsten Dierkes, Martin Bichler, and Ramayya Krishnan.
\newblock Estimating the effect of word of mouth on churn and cross-buying in
  the mobile phone market with markov logic networks.
\newblock {\em Decision Support Systems}, 51(3):361--371, 2011.

\bibitem[\protect\citeauthoryear{Fader \bgroup \em et al.\egroup
  }{2005}]{fader2005counting}
Peter~S Fader, Bruce~GS Hardie, and Ka~Lok Lee.
\newblock “counting your customers” the easy way: An alternative to the
  pareto/nbd model.
\newblock {\em Marketing science}, 24(2):275--284, 2005.

\bibitem[\protect\citeauthoryear{Glady \bgroup \em et al.\egroup
  }{2009}]{glady2009modeling}
Nicolas Glady, Bart Baesens, and Christophe Croux.
\newblock Modeling churn using customer lifetime value.
\newblock {\em European Journal of Operational Research}, 197(1):402--411,
  2009.

\bibitem[\protect\citeauthoryear{Hadden \bgroup \em et al.\egroup
  }{2007}]{hadden2007computer}
John Hadden, Ashutosh Tiwari, Rajkumar Roy, and Dymitr Ruta.
\newblock Computer assisted customer churn management: State-of-the-art and
  future trends.
\newblock {\em Computers \& Operations Research}, 34(10):2902--2917, 2007.

\bibitem[\protect\citeauthoryear{Handley}{1994}]{handley1994use}
Simon Handley.
\newblock On the use of a directed acyclic graph to represent a population of
  computer programs.
\newblock In {\em Proceedings of the First IEEE Conference on Evolutionary
  Computation. IEEE World Congress on Computational Intelligence}, pages
  154--159. IEEE, 1994.

\bibitem[\protect\citeauthoryear{Huang \bgroup \em et al.\egroup
  }{2018}]{huang2018improving}
Jin Huang, Wayne~Xin Zhao, Hongjian Dou, Ji-Rong Wen, and Edward~Y Chang.
\newblock Improving sequential recommendation with knowledge-enhanced memory
  networks.
\newblock In {\em The 41st International ACM SIGIR Conference on Research \&
  Development in Information Retrieval}, pages 505--514, 2018.

\bibitem[\protect\citeauthoryear{Jerath \bgroup \em et al.\egroup
  }{2011}]{jerath2011new}
Kinshuk Jerath, Peter~S Fader, and Bruce~GS Hardie.
\newblock New perspectives on customer “death” using a generalization of
  the pareto/nbd model.
\newblock {\em Marketing Science}, 30(5):866--880, 2011.

\bibitem[\protect\citeauthoryear{Ji}{2005}]{ji2005models}
Xiaoyu Ji.
\newblock Models and algorithm for stochastic shortest path problem.
\newblock {\em Applied Mathematics and Computation}, 170(1):503--514, 2005.

\bibitem[\protect\citeauthoryear{Karahoca and Karahoca}{2011}]{karahoca2011gsm}
Adem Karahoca and Dilek Karahoca.
\newblock Gsm churn management by using fuzzy c-means clustering and adaptive
  neuro fuzzy inference system.
\newblock {\em Expert Systems with Applications}, 38(3):1814--1822, 2011.

\bibitem[\protect\citeauthoryear{Kirui \bgroup \em et al.\egroup
  }{2013}]{kirui2013predicting}
Clement Kirui, Li~Hong, Wilson Cheruiyot, and Hillary Kirui.
\newblock Predicting customer churn in mobile telephony industry using
  probabilistic classifiers in data mining.
\newblock {\em International Journal of Computer Science Issues (IJCSI)}, 10(2
  Part 1):165, 2013.

\bibitem[\protect\citeauthoryear{Kolobov \bgroup \em et al.\egroup
  }{2011}]{kolobov2011heuristic}
Andrey Kolobov, Mausam Mausam, Daniel~S Weld, and Hector Geffner.
\newblock Heuristic search for generalized stochastic shortest path mdps.
\newblock In {\em Twenty-First International Conference on Automated Planning
  and Scheduling}, 2011.

\bibitem[\protect\citeauthoryear{Kumar \bgroup \em et al.\egroup
  }{2004}]{kumar2004customer}
V~Kumar, Girish Ramani, and Timothy Bohling.
\newblock Customer lifetime value approaches and best practice applications.
\newblock {\em Journal of interactive Marketing}, 18(3):60--72, 2004.

\bibitem[\protect\citeauthoryear{Ma and Qin}{2009}]{ma2009research}
Hongxia Ma and Min Qin.
\newblock Research method of customer churn crisis based on decision tree.
\newblock In {\em 2009 International Conference on Management and Service
  Science}, pages 1--4. IEEE, 2009.

\bibitem[\protect\citeauthoryear{Paquiens{\'e}guy and
  He}{2018}]{paquienseguy2018user}
Fran{\c{c}}oise Paquiens{\'e}guy and Miao He.
\newblock The user, as key element for platforms--through the lens of alibaba,
  2018.

\bibitem[\protect\citeauthoryear{Permana \bgroup \em et al.\egroup
  }{2014}]{permana2014study}
Dony Permana, Sapto~Wahyu Indratno, and Udjianna~S Pasaribu.
\newblock Study of behavior and determination of customer lifetime value (clv)
  using markov chain model.
\newblock In {\em AIP conference Proceedings}, volume 1589, pages 456--459.
  American Institute of Physics, 2014.

\bibitem[\protect\citeauthoryear{Shiau \bgroup \em et al.\egroup
  }{2018}]{shiau2018examining}
Wen-Lung Shiau, Yogesh~K Dwivedi, and He-Hong Lai.
\newblock Examining the core knowledge on facebook.
\newblock {\em International Journal of Information Management}, 43:52--63,
  2018.

\bibitem[\protect\citeauthoryear{Tsai and Chen}{2010}]{tsai2010variable}
Chih-Fong Tsai and Mao-Yuan Chen.
\newblock Variable selection by association rules for customer churn prediction
  of multimedia on demand.
\newblock {\em Expert Systems with Applications}, 37(3):2006--2015, 2010.

\bibitem[\protect\citeauthoryear{Zeithaml \bgroup \em et al.\egroup
  }{2001}]{zeithaml2001driving}
Valarie~A Zeithaml, Katherine~N Lemon, and Roland~T Rust.
\newblock {\em Driving customer equity: How customer lifetime value is
  reshaping corporate strategy}.
\newblock Simon and Schuster, 2001.

\bibitem[\protect\citeauthoryear{Zhao \bgroup \em et al.\egroup
  }{2020}]{zhao2020maximizing}
Yifei Zhao, Yu-Hang Zhou, Mingdong Ou, Huan Xu, and Nan Li.
\newblock Maximizing cumulative user engagement in sequential recommendation:
  An online optimization perspective.
\newblock In {\em Proceedings of the 26th ACM SIGKDD International Conference
  on Knowledge Discovery \& Data Mining}, pages 2784--2792, 2020.

\end{thebibliography}


\begin{thebibliography}{}

\bibitem[\protect\citeauthoryear{Abelson \bgroup \em et al.\egroup
  }{1985}]{abelson-et-al:scheme}
Harold Abelson, Gerald~Jay Sussman, and Julie Sussman.
\newblock {\em Structure and Interpretation of Computer Programs}.
\newblock MIT Press, Cambridge, Massachusetts, 1985.

\bibitem[\protect\citeauthoryear{Baumgartner \bgroup \em et al.\egroup
  }{2001}]{bgf:Lixto}
Robert Baumgartner, Georg Gottlob, and Sergio Flesca.
\newblock Visual information extraction with {Lixto}.
\newblock In {\em Proceedings of the 27th International Conference on Very
  Large Databases}, pages 119--128, Rome, Italy, September 2001. Morgan
  Kaufmann.

\bibitem[\protect\citeauthoryear{Brachman and
  Schmolze}{1985}]{brachman-schmolze:kl-one}
Ronald~J. Brachman and James~G. Schmolze.
\newblock An overview of the {KL-ONE} knowledge representation system.
\newblock {\em Cognitive Science}, 9(2):171--216, April--June 1985.

\bibitem[\protect\citeauthoryear{Gottlob \bgroup \em et al.\egroup
  }{2002}]{gls:hypertrees}
Georg Gottlob, Nicola Leone, and Francesco Scarcello.
\newblock Hypertree decompositions and tractable queries.
\newblock {\em Journal of Computer and System Sciences}, 64(3):579--627, May
  2002.

\bibitem[\protect\citeauthoryear{Gottlob}{1992}]{gottlob:nonmon}
Georg Gottlob.
\newblock Complexity results for nonmonotonic logics.
\newblock {\em Journal of Logic and Computation}, 2(3):397--425, June 1992.

\bibitem[\protect\citeauthoryear{Levesque}{1984a}]{levesque:functional-foundations}
Hector~J. Levesque.
\newblock Foundations of a functional approach to knowledge representation.
\newblock {\em Artificial Intelligence}, 23(2):155--212, July 1984.

\bibitem[\protect\citeauthoryear{Levesque}{1984b}]{levesque:belief}
Hector~J. Levesque.
\newblock A logic of implicit and explicit belief.
\newblock In {\em Proceedings of the Fourth National Conference on Artificial
  Intelligence}, pages 198--202, Austin, Texas, August 1984. American
  Association for Artificial Intelligence.

\bibitem[\protect\citeauthoryear{Nebel}{2000}]{nebel:jair-2000}
Bernhard Nebel.
\newblock On the compilability and expressive power of propositional planning
  formalisms.
\newblock {\em Journal of Artificial Intelligence Research}, 12:271--315, 2000.

\end{thebibliography}

\end{document}